\documentclass[lettersize,journal]{IEEEtran}
\usepackage{amsmath,amsfonts}
\usepackage{algorithmic}
\usepackage{algorithm}
\usepackage{amssymb}

\usepackage{array}
\usepackage[caption=false,font=footnotesize,labelfont=sf,textfont=sf]{subfig}
\usepackage{textcomp}
\usepackage{stfloats}
\usepackage{url}
\usepackage{verbatim}
\usepackage{graphicx}
\usepackage{cite}
\usepackage{booktabs}
\usepackage[font=small]{caption}
\usepackage[dvipsnames]{xcolor}
\usepackage{multirow}

\begin{document}

\title{Efficiently Learning Robust Torque-based Locomotion Through Reinforcement with Model-Based Supervision}

\author{Yashuai Yan*$^{,1}$, Tobias Egle*$^{,2}$, Christian Ott$^{2,3}$, and Dongheui Lee$^{1,3}$
	\thanks{$^{1}$Yashuai Yan and Dongheui Lee are with the Autonomous Systems Lab, TU Wien, 1040 Vienna, Austria
		{\tt\small \{yashuai.yan, dongheui.lee\}@tuwien.ac.at}}%
    \thanks{$^{2}$Tobias Egle and Christian Ott are with the Automation and Control Institute, TU Wien, 1040 Vienna, Austria
		{\tt\small \{tobias.egle, christian.ott\}@tuwien.ac.at}}%
	
  \thanks{$^{3}$Christian Ott and Dongheui Lee are also with the Institute of Robotics and Mechatronics (DLR), German Aerospace Center, Wessling, Germany.}%
    \thanks{*Contributed equally to the work.}
}

\maketitle

\begin{abstract}
We propose a control framework that integrates model-based bipedal locomotion with residual reinforcement learning (RL) to achieve robust and adaptive walking in the presence of real-world uncertainties. Our approach leverages a model-based controller—comprising a Divergent Component of Motion (DCM) trajectory planner and a whole-body controller—as a reliable base policy. To address the uncertainties of inaccurate dynamics modeling and sensor noise, we introduce a residual policy trained through RL with domain randomization. Crucially, we employ a model-based oracle policy, which has privileged access to ground-truth dynamics during training, to supervise the residual policy via a novel supervised loss. This supervision enables the policy to efficiently learn corrective behaviors that compensate for unmodeled effects without extensive reward shaping. Our method demonstrates improved robustness and generalization across a range of randomized conditions, offering a scalable solution for sim-to-real transfer in bipedal locomotion.
\end{abstract}


\section{Introduction}
\IEEEPARstart{L}{earning} effective torque-based locomotion policies for bipedal robots remains a central challenge in robotics. While torque control offers low-level access to a robot’s actuators—enabling responsive, energy-efficient, and compliant behaviors—it also demands precise handling of complex, nonlinear dynamics, and can lead to unstable behaviors if not carefully regulated, particularly in high-dimensional floating-base systems. This complexity makes direct policy learning at the torque level particularly difficult, especially in the presence of real-world factors such as contact uncertainty, sensor noise, and actuation delays.

Traditionally, bipedal locomotion has been addressed through model-based control methods \cite{kajita3DLinearInverted2001, sugiharaRealtimeHumanoidMotion2002, wieberTrajectoryFreeLinear2006, takenakaRealTimeMotion2009, englsbergerThreeDimensionalBipedalWalking2015}, which decompose the task into high-level trajectory planning and low-level whole-body control strategies. These controllers depend on accurate models of the robot’s dynamics and reliable state estimation to achieve stable walking motions. While model-based approaches often incorporate robustness to model uncertainties and external disturbances \cite{mesesanDynamicWalkingCompliant2019}, their performance can degrade when discrepancies between the model and the physical system become substantial, or when the state estimation is significantly affected by sensor noise. In practice, unmodeled dynamics such as joint friction, actuator nonlinearities, backlash, and structural flexibilities combined with noisy sensory data, can challenge control accuracy and consistency on the real robot.

An alternative strategy to address this challenge is iterative learning control. For example, Hu et al. \cite{7484896} proposed learning a compensatory Zero-Moment Point (ZMP) trajectory by observing and correcting ZMP errors over repeated executions. This approach refines the reference trajectory to mitigate the impact of unmodeled dynamics during the pattern generation stage. However, it relies on repetitive, consistent motion patterns to accumulate corrections, which limits its applicability in dynamic or highly variable tasks.  

To address these challenges, recent advances in deep reinforcement learning (DRL) \cite{rudin2022learning, xie2023learning, li2025reinforcement} have demonstrated promising results. By training policies with domain randomization in simulation, DRL methods \cite{10000225, xie2020learning, he2024learning, he2024omniho} can produce controllers that are robust to modeling errors and environmental variations. Nonetheless, purely learning-based approaches face limitations: they often require extensive data, are sensitive to reward design, and lack the interpretability and safety assurances of model-based methods.

To bridge this gap, recent works have explored hybrid strategies that combine the advantages of model-based and data-driven methods. For instance, Duan et al. \cite{duanLearningTaskSpace2021} integrated the knowledge of the robot system into DRL to learn a task-space policy rather than joint-level controllers. This high-level policy is then connected with a low-level inverse dynamics controller to command joint torques to the robot. Similarly, Castillo et al. \cite{castillo2023template} incorporated insights from an angular momentum-based linear inverted pendulum model and designed a hierarchical framework in which the RL learns to generate high-level trajectories, followed by a low-level task-space tracking controller. Besides task-space learning, Egle et al. \cite{10769960} applied RL to learn step location and timing adaptation to enhance the robustness of model-based controllers against strong external disturbances. While these approaches reduce reward engineering and improve sample efficiency, they still rely solely on reward signals for policy learning—requiring a number of trial-error iterations and offering limited guidance during training.

In contrast to standard approaches that rely solely on reward shaping, we propose a supervised reinforcement learning framework that directly incorporates an additional supervised loss term into the training objective. Our control architecture combines the structured reliability of model-based controllers with the adaptability of reinforcement learning. Specifically, we augment a model-based bipedal locomotion controller with a residual RL policy that learns to compensate for model inaccuracies. To expose the policy to real-world uncertainties during training, we propose to apply dynamics randomization and introduce a privileged Oracle policy that has full access to ground-truth parameters and generates ideal corrective actions under randomized conditions. Rather than learning solely from sparse or shaped rewards, the RL policy is explicitly guided by the Oracle through the supervised loss, which enhances learning efficiency and convergence speed. By integrating model-based priors with data-driven adaptation, our approach improves efficiency, robustness, and generalization across different bipedal robot platforms.

We summarize the key contributions of this work as follows:

\begin{itemize}
\item A control framework that integrates model-based control with data-driven reinforcement learning to learn residual corrective actions for mitigating model inaccuracies.

\item A model-based oracle policy that leverages privileged information to compensate for system uncertainties and generate near-optimal control signals, which are used to supervise the learning process.

\item A supervised reinforcement learning framework that incorporates Oracle actions into a supervised loss, complementing the standard RL objective and improving learning efficiency without relying exclusively on reward shaping.

\item Comprehensive evaluations on three bipedal robots: Kangaroo, Unitree H1-2, and Bruce, demonstrating the robustness and generalization capabilities of our approach across different platforms.

\end{itemize}

\section{Methodology}

\begin{figure*}[]
    \centering
    \includegraphics[width=0.9\textwidth]{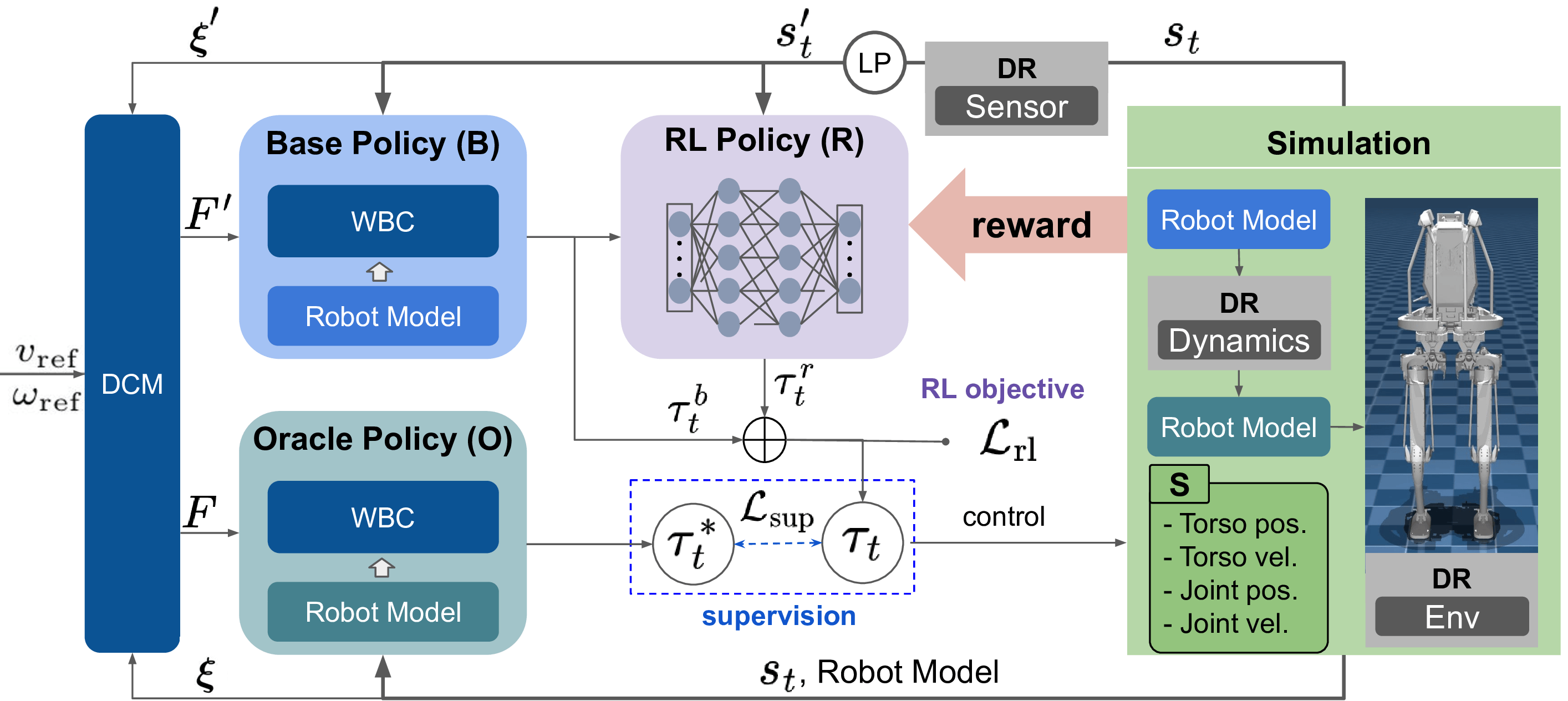}
    \caption{\textbf{Residual Reinforcement Learning with Model-Based Supervision.} Our framework integrates three key components: the Base Policy, the Oracle Policy, and the Residual Policy. The Base and Oracle policies are model-based controllers; however, the Oracle, only used in training, has privileged access to true system information, including the robot model, state, and motor parameters, while the Base Policy operates under an inaccurate dynamics model with realistic assumptions without access to the true system information (LP: low-pass filter). The learnable Residual Policy is trained to compensate for the model inaccuracies of the Base Policy. Crucially, the Residual Policy is guided by both the RL objective $\mathcal{L}_{\text{rl}}$ and direct supervision from the Oracle Policy $\mathcal{L}_{\text{sup}}$, enabling efficient learning and improved robustness under real-world uncertainties.}
    \label{fig:modeloverview}
\end{figure*}
In this section, we introduce our method, which integrates the strengths of both reinforcement learning and model-based robot control algorithms. We begin by describing the model-based algorithms that serve as the base policy. Next, we present a reinforcement learning-based residual policy that enhances the robustness of the model-based approach, particularly in addressing inaccuracies arising from unrealistic robot modeling in real-world scenarios. An overview of our method is provided in Figure \ref{fig:modeloverview}.

\subsection{Model-based Robot Control}
Our model-based bipedal locomotion controller is composed of two key components: a trajectory generator based on the Divergent Component of Motion (DCM) and an inverse dynamics-based whole-body controller (WBC).
\subsubsection{DCM Trajectory Generation}
The trajectory generation framework leverages the concept of three-dimensional DCM and the Virtual Repellent Point (VRP) \cite{englsbergerThreeDimensionalBipedalWalking2015, 10288271}. The DCM extends the notion of the capture point (CP) into three dimensions and is defined as:
\begin{equation}
    \boldsymbol{\xi} = \boldsymbol{x} + b \boldsymbol{\dot{x}}.
\end{equation}
where $\boldsymbol{x}$ and $\boldsymbol{\dot{x}}$ represent the Center of Mass (CoM) position and velocity, respectively. $b=\sqrt{\Delta z / g}$ is derived from the average CoM height $\Delta z$ and gravitational acceleration $g$. From the CoM dynamics $\boldsymbol{\ddot{x}} = \boldsymbol{g} + \boldsymbol{F}_{\text{ext}}/m$, we derive the DCM dynamics as:
\begin{equation}
    \boldsymbol{\dot{\xi}} = \frac{1}{b} (\boldsymbol{\xi} - \boldsymbol{v}),
\label{eq:dcm_dynamics}
\end{equation}
where $\boldsymbol{v} = \boldsymbol{x} - b^2/m \boldsymbol{F}$  denotes the VRP, which encodes the total force acting on the CoM, i.e., $\boldsymbol{F} = m\boldsymbol{g} + \boldsymbol{F}_{\text{ext}}$.

To generate walking trajectories, we plan a sequence of $n$ preview VRP waypoints $\{\boldsymbol{v}_1, \cdots, \boldsymbol{v}_n\}$, which define $n-1$ transition phases alternating between single and double support phases. Within each phase $\psi$, over a time interval $t \in [0, T]$, the DCM trajectory is determined by solving (\ref{eq:dcm_dynamics}) with a terminal constraint $\boldsymbol{\xi}_{\psi}(T)$. The solution is given by:
\begin{equation}
    \boldsymbol{\xi}_{\psi}(t) = \alpha_{\psi}(t)\boldsymbol{v}_{\psi, 0} + \beta_{\psi}(t) \boldsymbol{v}_{\psi, T} + \gamma_{\psi}(t) \boldsymbol{\xi}_{\psi}(T),
\end{equation}
where the nonlinear coefficients $\alpha_{\psi}(t), \beta_{\psi}(t)$ and $\gamma_{\psi}(t)$ depend on the trajectory of the VRP from the start point $\boldsymbol{v}_{\psi, 0}$ to the endpoint $\boldsymbol{v}_{\psi, T}$ \cite{mesesanConvexPropertiesCenterofMass2018}. We define equality constraints for the start and end points of adjacent transition phases to ensure the continuity of the trajectory. The complete trajectory is obtained by starting from a DCM endpoint and computing backward in time. 

In addition to the DCM trajectory, our model-based method also plans the continuous foot trajectories. Similarly to prior work \cite{10000236}, we apply six-order polynomial functions to parameterize the foot trajectories by conditioning on the planned foot locations.

\subsubsection{Whole-body Robot Control}
To track the planned reference trajectories, we employ an inverse dynamics-based WBC \cite{englsbergerTorqueBasedDynamicWalking2018a}. A key objective of the controller is to stabilize the inherently unstable first-order DCM dynamics in (\ref{eq:dcm_dynamics}). Therefore, we implement the following control law \cite{englsbergerThreeDimensionalBipedalWalking2015}:
\begin{equation}\label{eq:DCMcontroller}
\boldsymbol{v} = \boldsymbol{v}_{\text{ref}} + (\boldsymbol{I} + b \mathbf{K}_{\xi})(\boldsymbol{\xi} - \boldsymbol{\xi}_{\text{ref}}),
\end{equation}
where $\boldsymbol{v}_{\text{ref}}$ is the reference VRP, $\mathbf{K}_{\xi}$ is a positive definite diagonal gain matrix, and $\boldsymbol{e}_{\xi} = \boldsymbol{\xi} - \boldsymbol{\xi}_{\text{ref}}$ is defined as the DCM tracking error. From the output of the DCM controller (\ref{eq:DCMcontroller}) we can compute the desired force on the CoM as
\begin{equation}
    \boldsymbol{F} = m/b^2 (\boldsymbol{x}- \boldsymbol{v}),
\end{equation}
which is commanded as a reference in the CoM task of the whole-body controller. Other tasks include foot trajectory tracking, reference tracking in joint or task space for the arms (if available), and maintaining a desired body orientation. Further details on the WBC formulation can be found in \cite{englsbergerTorqueBasedDynamicWalking2018a}.

\subsection{Enhance Robustness through Randomization}

\begin{table}[]
\vspace*{5pt}
\centering
\resizebox{0.95\linewidth}{!}{%
\begin{tabular}{lccc}
Parameter & Unit & Distribution & Operator \\ \midrule
\multicolumn{4}{c}{\textbf{Sensory Noise}}  \\ \midrule
Torso position &  m &  $\mathcal{N}(0, 0.05\beta)$ & additive \\
Torso rotation &  rad & $\mathcal{N}(0, 0.05\beta)$  & additive \\
Linear velocity & m/s  & $\mathcal{N}(0, 0.1\beta)$ & additive \\
Angular velocity & rad/s  & $\mathcal{N}(0, 0.1\beta)$ & additive \\
Joint position & rad  & $\mathcal{N}(0, 0.05\beta)$ & additive \\
Joint velocity & rad/s & $\mathcal{N}(0, 0.1\beta)$ & additive \\ \midrule
\multicolumn{4}{c}{\textbf{Dynamics Uncertainty}}  \\ \midrule
Body mass & - & $\mathcal{U}(1-0.2\beta, 1+0.5\beta)$ & scaling \\
Joint friction & - & $\mathcal{U}(1-0.5\beta, 1+0.1\beta)$ & scaling  \\
Joint damping & - & $\mathcal{U}(1-0.5\beta, 1+0.2\beta)$ & scaling  \\ 
Motor efficiency $\alpha_{\text{decay}}$ & - & $\mathcal{U}(1-0.2\beta, 1.0)$ & scaling  \\
Motor delay $\Delta t_{delay}$ & ms & $\mathcal{U}(2, 4)$ & additive  \\ \midrule
\multicolumn{4}{c}{\textbf{Environment}}  \\ \midrule
Floor friction & - & $\mathcal{U}(0.5, 1.1)$ & scaling  \\
\end{tabular}
}
\caption{\textbf{Domain Randomization.} We capture variability from different sources and use a scaling factor $\beta$ to control the randomization level during validation. For training, we use $\beta=1$. }
\label{tab:domain_randomization}
\end{table}

To address real-world variability, our method employs domain randomization (DR) during reinforcement learning, targeting three principal sources of uncertainty: sensors, robot dynamics, and environment. Sensor noise is simulated by perturbing the robot state inputs to the model-based controller. Dynamics variability is introduced by randomizing the physical properties of the robot, such as mass, inertia, motor strength, and actuation delay. Since detailed information on actuators and transmissions is not available for all platforms, we apply randomization at the joint level. The randomization ranges are chosen to cover plausible variations induced by the underlying actuators. Additionally, we vary ground friction coefficients and include ground unevenness to reflect the diversity of real-world terrain conditions. A comprehensive list of the randomized parameters used in our simulation environment is provided in Table \ref{tab:domain_randomization}.

\subsection{Model-based Supervision}

\begin{algorithm}[t]
\caption{Learning Procedure}
\label{alg:oracle}
\begin{algorithmic}[1]
\STATE \textbf{Init:} RB, $M$, $\pi_b$, $\pi_*$, $\pi_r$, $\Delta t_{\text{sim}}$
\WHILE{$\pi_r$ not converged}
\FOR{each episode}
    \STATE $M' \leftarrow DR(M)$, set simulator with $M'$
    \STATE $s'_0 \leftarrow DR(s_0)$
    \STATE sample $\Delta t_{\text{delay}}, \alpha_{\text{decay}}$
    \FOR{each step $t$}
        \STATE $\tau^b_t \leftarrow \pi_b(s'_t)$, $\tau^r_t \leftarrow \pi_r(s'_t)$
        \STATE $\tau_t \leftarrow \tau^b_t + \tau^r_t$
        \STATE simulate $\Delta t_{\text{sim}}-\Delta t_{\text{delay}}$ with $\alpha_{\text{decay}}\tau_t$
        \STATE $s'_{t+\Delta t_{\text{sim}}} \leftarrow DR(s_{t+\Delta t_{\text{sim}}-\Delta t_{\text{delay}}})$
        \STATE $\tau^b_{t+\Delta t_{\text{sim}}} \leftarrow \pi_b(M,s'_{t+\Delta t_{\text{sim}}-\Delta t_{\text{delay}}})$
        \STATE simulate $\Delta t_{\text{delay}}$ with $\alpha_{\text{decay}}\tau_t$
        \STATE sample $\Delta t_{\text{delay}}, \alpha_{\text{decay}}$
        \STATE $\tau^*_{t+1} \leftarrow \pi_*(M',s_{t+\Delta t_{\text{sim}}})/\alpha_{\text{decay}}$
        \STATE $R_t \leftarrow$ reward
        \STATE RB $\leftarrow (s'_t,\tau^b_{t+1},\tau^*_{t+1},\tau^r_t,R_t,s'_{t+1})$
    \ENDFOR
    \FOR{each update}
        \STATE sample $\mathcal{B}$ from RB
        \STATE compute $\mathcal{L}_{\text{rl}}, \mathcal{L}_{\text{sup}}$
        \STATE update policy and critic
    \ENDFOR
\ENDFOR
\ENDWHILE
\end{algorithmic}
\end{algorithm}
Our framework incorporates the strength of model-based methods by explicitly utilizing knowledge of system uncertainties—information that is accessible with known randomization. To realize this, we introduce three key components: a Base policy, an Oracle policy, and a residual RL policy. The Base and Oracle policies are model-based controllers, while the residual policy learns to control robots with RL algorithms by interacting with the environment.

The Base policy, denoted as $\pi_b$, operates without access to the underlying randomization parameters. It operates on noisy state observations and uses an inaccurate robot model. In contrast, the Oracle policy, denoted as $\pi_{*}$, has complete information on the ground-truth randomization settings and the exact robot model within the simulator. This privileged information enables $\pi_{*}$ to model uncertainties. For example, $\pi_{*}$ has access to the real signal without noise and computes torques based on accurate robot dynamics. Additionally, $\pi_{*}$ is aware of variations in motor characteristics—actuation delays $\Delta t_{\text{delay}}$ and motor efficiency $\alpha_{\text{decay}}$—and incorporates this knowledge when computing torque commands. A detailed learning procedure is illustrated in Algorithm \ref{alg:oracle}. 

The oracle policy $\pi_{*}$ thus serves as an idealized expert supervisor, leveraging its privileged access to ground-truth dynamics and system parameters to generate ideal control signals. Its role is to guide the residual policy in compensating for the limitations of the Base policy $\pi_b$, which arise from unmodeled dynamics and unknown variability. By learning to mimic the corrections provided by $\pi_{*}$, the residual policy enhances the overall control system’s robustness and adaptability to real-world uncertainties.
    
\subsection{Supervised Reinforcement Learning}
To improve the robustness and adaptability of model-based control in the presence of real-world uncertainties, we formulate a residual RL problem. In this framework, the residual policy learns corrective actions on top of a model-based Base policy through interaction with a randomized simulation environment. The residual learning allows the agent to adapt to dynamics and disturbances that are difficult to model explicitly. In the following, we describe the design of the observation space and reward functions used to train the residual policy.
\subsubsection{Observation Space}
To enable the residual policy to infer latent dynamics and compensate for unmodeled variability, we design an observation space that captures both the instantaneous robot state and its temporal evolution. Specifically, we employ a recurrent neural network (RNN) to process the observation history $\boldsymbol{O}_{1:t} = \left[ \boldsymbol{o}_1, \boldsymbol{o}_2, \cdots, \boldsymbol{o}_t \right]$, allowing the policy to leverage temporal patterns for more informed decision-making. At each timestep $t$, the observation vector $\boldsymbol{o}_t$ is defined as follows:
\begin{equation}
    \boldsymbol{o}_t = \left[\boldsymbol{q}'_t, \boldsymbol{\dot{q}}'_t, \boldsymbol{\tau}^b_t, \boldsymbol{\tau}_{t-1}, \boldsymbol{e}'_{\xi, t}, \boldsymbol{e}'_{\text{foot}, t} \right],
\end{equation}
where $\boldsymbol{q}'_t$ and $\boldsymbol{\dot{q}}'_t$ denote the joint position and velocity after randomization, $\boldsymbol{\tau}^b_t$ is the torque output of the base policy, $\boldsymbol{\tau}_{t-1}$ is the last action, and $\boldsymbol{e}'_{\xi, t}$ and $\boldsymbol{e}'_{\text{foot}, t}$ are the tracking errors of the DCM and foot trajectories from the Base policy. This observation reflects the noisy and uncertain sensory input available in real-world scenarios. 

To provide richer observation during training, we also define a privileged observation used exclusively by the critic network. The privileged observation vector is given by: 
\begin{equation}
    \boldsymbol{o}^{\text{privi.}}_{t} = \left[ \boldsymbol{o}_t, \boldsymbol{v}_t, \boldsymbol{\omega}_t, \boldsymbol{q}_t, \boldsymbol{\dot{q}}_t, \boldsymbol{\tau}^{*}_t, \boldsymbol{e}_{\xi, t}, \boldsymbol{e}_{\text{foot}, t} \right],
\end{equation}
where $\boldsymbol{v}_t$ and $\boldsymbol{\omega}_t$ are linear and angular velocity of the floating base. $\boldsymbol{q}_t$ and $\boldsymbol{\dot{q}}_t$ are the true joint states, and $\boldsymbol{\tau}^{*}_t$ is the torque computed by the oracle policy $\pi_{*}$. Besides, the critic has access to the tracking errors $\boldsymbol{e}_{\xi, t}, \boldsymbol{e}_{\text{foot}, t}$ from the Oracle policy. This privileged input enables more accurate value estimation during training, without being available to the policy during deployment.

\subsubsection{Reward Functions}

\begin{table}[]
\centering
\vspace*{5pt}
\resizebox{0.9\linewidth}{!}{%
\begin{tabular}{lccc}
Reward & Expression        & Distance & Parameter ($w, \lambda$) \\ \midrule
\multicolumn{4}{c}{\textbf{Tracking}}                      \\  \midrule
$R_{\xi}$      & \multirow{4}{*}{$w\exp{(-\lambda d)}$} &  $d=||\boldsymbol{e}_{\xi}||_2$   & $(20, 10)$     \\
$R_{\text{foot}}$       &   &      $d=||\boldsymbol{e}_{\text{foot}}||_2$    &     $(5, 10)$      \\
$R^{\text{rot}}_{\text{torso}}$    & &    $d=||\boldsymbol{e}^{\text{rot}}_{\text{torso}}||_2$  & $(1, 10)$ \\
$R_{\tau}$       &  &   $d=||\boldsymbol{\tau} - \boldsymbol{\tau}^{*}||_2$       &   $(5, 0.01)$     \\  \midrule
\multicolumn{4}{c}{\textbf{Regularization}}   \\   \midrule
 $R_{\text{smooth}}$      &  &     $d=||\boldsymbol{\tau}_{t+1} - \boldsymbol{\tau}_{t}||_2$     &    $(0.01, 0.01)$       \\  \midrule
\multicolumn{4}{c}{\textbf{Punishment}}                        \\  \midrule
$R_{\text{termination}}$            & \multicolumn{3}{c}{-20 if early terminated; 0 otherwise}                       
\end{tabular}
}
\caption{\textbf{Reward Functions.}}
\label{tab:reward}
\end{table}

The reward functions used in this work are summarized in Table~\ref{tab:reward}. The primary components are tracking rewards that encourage the RL policy to follow the planned DCM and foot trajectories accurately. The torque tracking reward $R_{\tau}$ guides the applied torque $\boldsymbol{\tau} = \boldsymbol{\tau}^b + \boldsymbol{\tau}^r$ to match the output of the oracle policy $\boldsymbol{\tau}^{*}$, promoting consistency with optimal control behavior.

Additionally, $R_{\text{smooth}}$ is included to promote smooth transitions in the control signals. Early termination is triggered when the DCM error $||\boldsymbol{e}_{\xi}||_2$ exceeds 0.2 meters, and this event is penalized through the $R_{\text{termination}}$ term.

Notably, our framework benefits from supervised signals provided by the oracle policy during training, which reduces reliance on extensive reward engineering and hyperparameter tuning typically required in conventional RL approaches.

\subsection{Supervised Loss}
To train the residual policy, we adopt the Proximal Policy Optimization (PPO) algorithm~\cite{schulman2017proximal} with the objective $\mathcal{L}_{\text{rl}}$. Meanwhile, we leverage supervision from the Oracle policy and introduce an additional supervised loss term $\mathcal{L}_{\text{sup}}$ that encourages the policy to assign a higher likelihood to the oracle-corrected residual action. Specifically, the supervised loss is formulated as:
\begin{equation}
\mathcal{L}_{\text{sup}} = -\log (\pi_{r}(\boldsymbol{\tau}^{*}_t - \boldsymbol{\tau}^{b}_t| s_t)),
\label{eq:supervised_loss}
\end{equation}
where $\boldsymbol{\tau}^{*}_t$ is the oracle torque and $\boldsymbol{\tau}^{b}_t$ is the base policy torque. This loss guides the policy toward imitating the oracle's corrective behavior in response to system uncertainties.

The total training objective combines the PPO loss and the supervised loss:
\begin{equation}
    \mathcal{L}_{total} = \omega_{\text{rl}} * \mathcal{L}_{\text{rl}} + \omega_{\text{sup}} \mathcal{L}_{\text{sup}},
    \label{eq:total_loss}
\end{equation}
where the weights $\omega_{\text{rl}}=1, \omega_{\text{sup}}=10$ balance the trade-off between imitation of the oracle policy and autonomous policy exploration.

\section{Evaluation}
We conduct comprehensive simulations on the different robots to evaluate the effectiveness of our proposed method. Specifically, we validate the following key questions:

\begin{itemize}
    \item \textbf{Q1}: Can our Oracle policy serve as an effective supervisory signal under extensive randomization of robot dynamics and system-level uncertainties?
    \item \textbf{Q2}: Can our residual policy learn to compensate for the unmodeled variability that the base policy fails to address?
    \item \textbf{Q3}: Can our approach outperform standard RL methods that rely solely on reward signals or imitation learning that only learns from supervision?
    \item \textbf{Q4}: Can our framework be applied across different platforms without any parameter tuning?
\end{itemize}

\subsection{Technical Implementation}
In our actor-critic framework, the actor network is composed of a two-layer LSTM module followed by a linear output layer. Each LSTM layer contains 256 hidden units, and the final output layer has 512 neurons. The critic network is implemented using a multi-layer perceptron (MLP) with a 512-neuron input layer followed by two hidden layers of 256 neurons each. We employ Exponential Linear Units (ELU) \cite{clevert2016fast} as the activation function for all hidden layers.


\begin{table*}[]
\vspace*{5pt}
\centering
\resizebox{0.99\textwidth}{!}{%
\begin{tabular}{lcccccccccccc}
\multicolumn{1}{c}{} & \multicolumn{3}{c}{Success Rate $\uparrow$ (in \%)} & 
\multicolumn{3}{c}{DCM Tracking $\downarrow$ (in cm)} & \multicolumn{3}{c}{Foot Tracking $\downarrow$ (in cm)} 
& \multicolumn{3}{c}{Return $\uparrow$ ($\times 10^3$)} \\ \cmidrule(lr){2-4} \cmidrule(lr){5-7} \cmidrule(lr){8-10} \cmidrule(lr){11-13}
& Kangaroo & Bruce & H1-2 & Kangaroo & Bruce & H1-2 & Kangaroo & Bruce & H1-2 & Kangaroo & Bruce & H1-2 \\ \cmidrule(lr){2-4} \cmidrule(lr){5-7} \cmidrule(lr){8-10} \cmidrule(lr){11-13}
Oracle (O)    & \textbf{100.0} & \textbf{100.0} & \textbf{100.0} & $3.05\pm0.30$ & $\boldsymbol{1.49\pm0.07}$ & $\boldsymbol{2.26\pm0.39}$ & $\boldsymbol{0.00\pm 0.00}$ & $\boldsymbol{0.00\pm 0.00}$ & $\boldsymbol{0.00\pm 0.00}$ &  \textbf{27.86} & \underline{28.16} &  \textbf{30.09}  \\ 

BasePolicy (B)  & 0.0 & 0.0 & 0.0 & $12.73\pm1.23$ & $8.33\pm1.66$ & $12.42\pm1.53$ & $0.01\pm0.01$ & $0.02\pm0.01$ & $0.04\pm0.03$ & 2.40 & 2.53 &  2.56  \\ \midrule

ResRL (BR)  &  10.0 & 0.0 & 0.0  &  $14.47 \pm 1.39$ & $10.18\pm1.45$ &  $11.93\pm 1.43$  & $69.51 \pm 0.40$ & $3.99\pm 3.22$ & $7.61\pm5.61$ &  1.72  &   0.89    &  0.20  \\ 
IL & \textbf{100.0} & 70.0 & 80.0 & $3.86 \pm 0.88 $ & $3.44 \pm 1.30$ & $3.65 \pm 1.68$  & $0.30 \pm 0.04$  & $0.70 \pm 0.63$ & $0.62 \pm 0.67$  & 24.9 & 23.45 &  25.96 \\
MBC &  0.0  & 0.0 &  0.0  &  $23.45\pm 2.45$  &  $16.14\pm 2.06$ &  $22.01\pm 2.19$  & $1.42\pm 0.74$  & $3.05\pm 1.69$ &  $2.21\pm 0.94$  &  0.72   & 0.45 &  0.63  \\ \midrule

Ours (OR)     & 96.7 & 80.0 & \textbf{100.0} & $4.64\pm 1.85$ & $2.27\pm0.78$ & $\underline{2.32\pm 0.24}$ & $0.10\pm 0.02$ & $0.54\pm0.08$ & $\underline{0.27\pm 0.04}$ & 21.94 & 26.17 & \underline{26.45} \\ 

Ours (BOR) & \textbf{100.0} & \textbf{100.0}  & \textbf{100.0} & $\boldsymbol{2.61 \pm 0.39}$ & $\underline{1.57\pm0.16}$ & $2.39\pm 0.32$ & $\underline{0.08\pm 0.01}$ & $\underline{0.45\pm 0.04}$ & $0.28\pm 0.08$ & \underline{25.74} & \textbf{28.22} & \underline{26.45} \\
\end{tabular}%
}
\caption{\textbf{Comparison of different approaches}. All methods are evaluated with domain randomization ($\beta=1$). The Oracle and Base in the first section indicate our Oracle and Base policies. The second section presents the performance of our baselines. We showcase the benefits of the supervision from our Oracle policy in the last section via the methods OR and BOR. OR indicates the the RL policy learns the direct troque commands, while BOR learns the residual torques of the Base policy.}
\label{tab:comparison}
\end{table*}

\begin{table}[]
\centering
\begin{tabular}{lcccccccccccc}
         & \multicolumn{2}{c}{Success Rate} & \multicolumn{2}{c}{DCM Tracking} & \multicolumn{2}{c}{Foot Tracking} \\ \cmidrule(lr){2-3} \cmidrule(lr){4-5} \cmidrule(lr){6-7} 
level $\beta$ & MBC & BOR & MBC & BOR & MBC & BOR   \\ \midrule
$0.1$ & \textbf{100.0} & \textbf{100.0} & \textbf{3.55} & 4.74 & 0.15 & \textbf{0.08} \\
$0.3$ & 90.0 & \textbf{100.0} & 7.09 & \textbf{3.94} & 0.54 &  \textbf{0.08} \\
$0.5$ & 50.0 & \textbf{100.0} & 13.35 & \textbf{3.33} & 1.08 &  \textbf{0.07}  \\
$0.7$ & 10.0 & \textbf{100.0} & 20.37 & \textbf{2.98} & 1.45 &  \textbf{0.07}
\end{tabular}%
\caption{\textbf{Randomization level analysis.} During the inference, we evaluate different domain randomization on Kangaroo by adjusting the parameter $\beta$, and compare their impacts on MBC and our method BOR. Larger $\beta$ indicates more uncertainties in the system.}
\label{tab:noise_level}
\end{table}

We train and evaluate our framework on three bipedal robots: Kangaroo \cite{roigHardwareDesignControl2022}, Unitree H1-2, and Bruce\cite{liuDesignControlMiniature2022}. The details of the robots are illustrated in Table \ref{tab:robots}. MuJoCo \cite{todorov2012mujoco} is used as the physics simulator to simulate the robots. We use PIQP \cite{schwan2023piqp} to solve the QP for the whole-body controller, achieving a computation time of approximately 1 ms. While the simulation runs at 1000 Hz, control commands are applied at a frequency of 200 Hz. The RL policies are trained over $10k$ episodes with each episode corresponding to 10 seconds of simulation.

\subsection{Baselines}
\label{sec:baselines}
\subsubsection{Model-based Controller (MBC)}
We compare our method with a model-based controller. For a fair comparison, we apply the following steps to handle the unmodeled uncertainties in MBC:
\begin{itemize}
    \item using a low-pass filter on robot state observations to mitigate the sensor noise;
    \item operating at a frequency of $1000 Hz$.
\end{itemize}

\subsubsection{Residual RL (ResRL)}
To enable a fair comparison with standard RL settings, we train an additional residual policy under the same conditions but without the supervised loss term in Eq.~\ref{eq:total_loss} (i.e., $(\omega_{\text{sup}}=0)$). However, we retain the torque-tracking reward ($R_\tau$), which continues to encourage the learned policy to follow the Oracle. This setup allows us to isolate and examine the impact of using an implicit reward signal versus an explicit loss formulation for incorporating Oracle guidance into the policy. For the same reason, we adopt a torque-based residual learning policy as our baseline, rather than the classical residual RL framework that learns joint positions with an underlying PD controller.


\subsubsection{Imitation Learning (IL)}
Since our method shares the similar idea as imitation learning, in which the student (Base+RL policy) learns to mimic the teacher (Oracle policy) behavior, we exploit our framework for imitation learning and compare it with our method. For IL in our framework, we skip the RL objective (i.e. $\omega_{\text{RL}} = 0$ in Eq. \ref{eq:total_loss}) and only optimize the policy network based on supervision signal.

\begin{table}[]
\centering
\begin{tabular}{l|cccc}
         & Total Mass & Height & DoFs per Leg & Foot \\ \cmidrule(lr){1-5}
Kangaroo & 47 Kg      & 145cm  & 6            & flat \\
Unitree H1-2       & 67Kg       & 178cm  & 6            & flat \\
Bruce    & 4.8Kg      & 70cm   & 5            & line
\end{tabular}%

\caption{\textbf{Properties of different robots}.}
\label{tab:robots}
\end{table}
\subsection{Quantitative and Qualitative Evaluation}
\subsubsection{Metrics}
\label{sec:metrics}

To compare with the baseline methods, we define the following evaluation metrics:
\begin{itemize}
    \item \textbf{Success Rate}: measures whether the robot can walk successfully for 5 seconds while following random velocity commands. A trial is considered a failure if the robot falls.
    \item \textbf{DCM Tracking}: assesses the distance between the measured and desired DCM positions. Accurate DCM tracking is critical for stable walking; large DCM errors can lead to instability or falls. 
    \item \textbf{Foot Tracking}: evaluates the accuracy of foot trajectory tracking, which reflects the robot’s ability to follow the commanded velocity inputs.
    \item \textbf{Return}: represents the cumulative reward as defined in Table~\ref{tab:reward}, excluding the torque tracking reward $R_\tau$ to avoid biasing the evaluation in favor of the oracle policy. This metric captures overall controller performance, including aspects such as trajectory tracking accuracy and motion smoothness.
\end{itemize}

\begin{figure*}
\vspace*{5pt}
    \centering
    \includegraphics[width=0.99\textwidth]{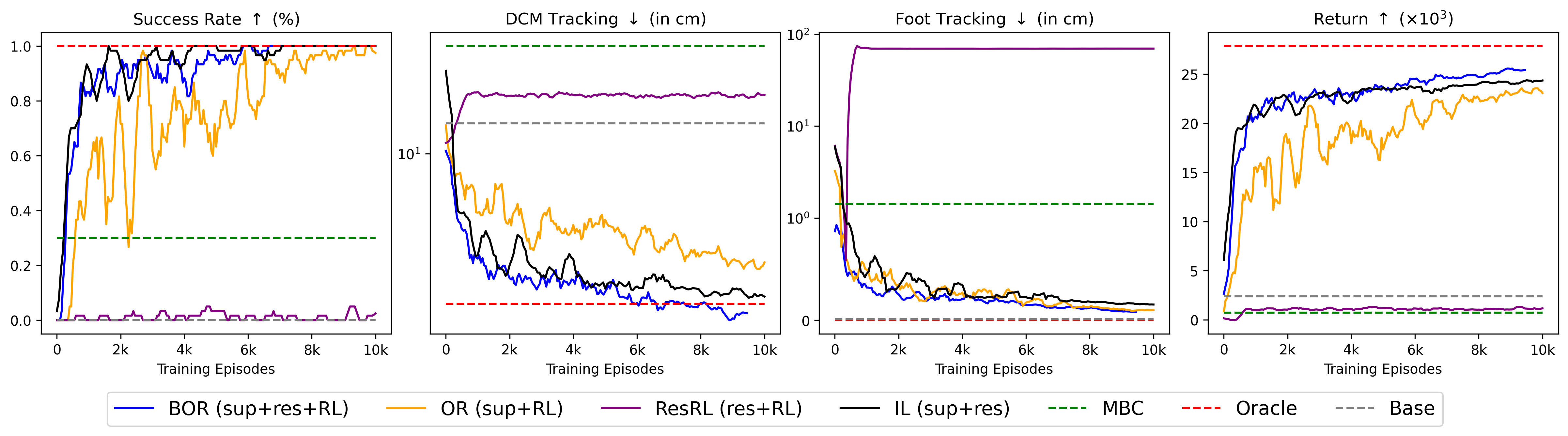}
    \caption{\textbf{Quantitative evaluation on the Kangaroo robot.}We evaluate various methods using predefined metrics throughout training. Our method (BOR) shows that residual learning converges substantially faster than directly learning torque commands, even under identical Oracle supervision. Moreover, comparing BOR/OR with other baselines demonstrates that incorporating the supervision term into the optimization objective significantly improves training efficiency, bringing performance much closer to that of the Oracle policies.}
    \label{fig:training_procedure}
\end{figure*}

\begin{figure}
    \centering
    \includegraphics[width=0.95\linewidth]{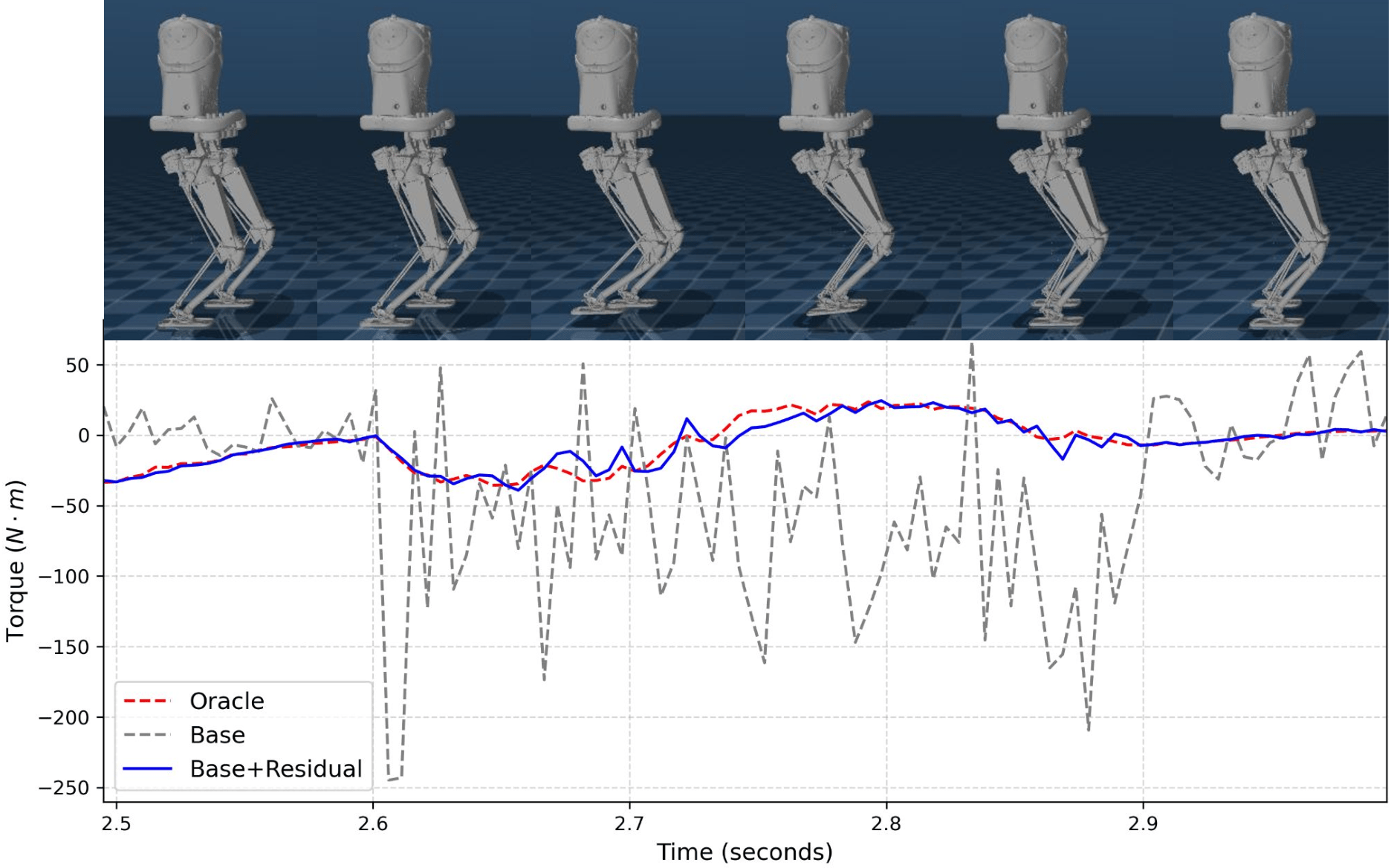}
    \caption{\textbf{Torque tracking on the right hip pitch joint.} The Base policy produces noisy torques due to the domain randomization. Our learned residual policy succeeds in compensating the joint torques, closely tracking the Oracle policy.}
    \label{fig:torque_tracking}
\end{figure}

\begin{figure*}
    \centering
    \includegraphics[width=0.95\linewidth]{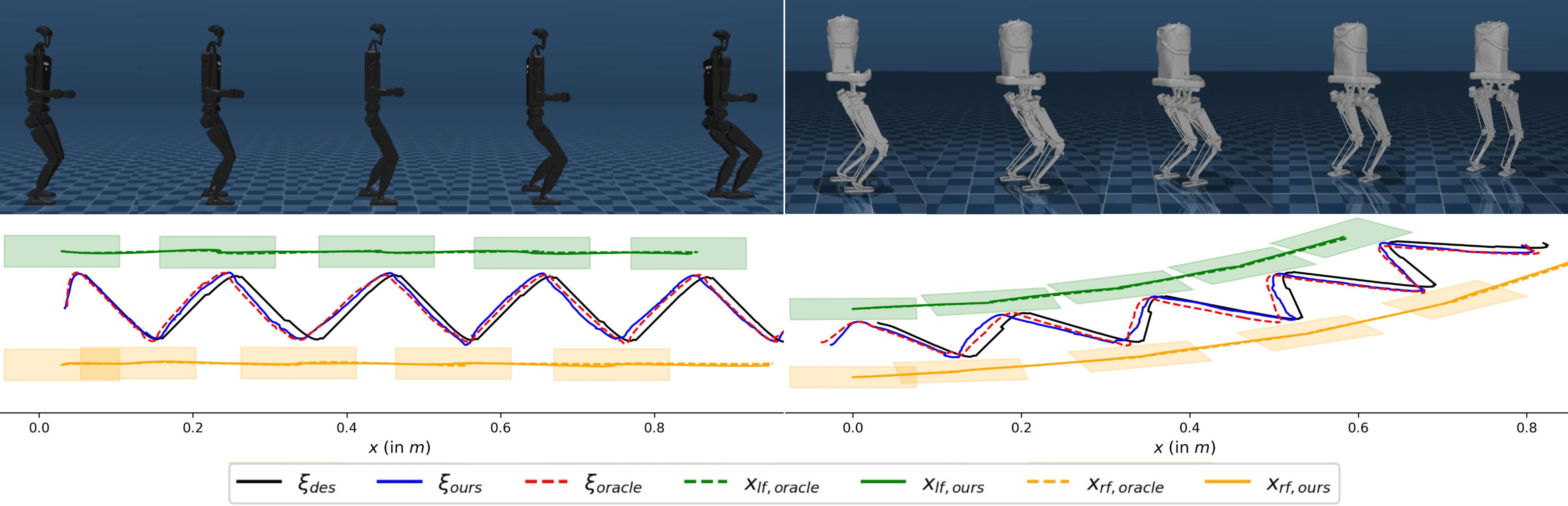}
    \caption{\textbf{Quantitative evaluation on DCM and foot tracking.} The H1-2 robot walks straight forward at a speed of $0.2 m/s$, while the Kangaroo is commanded with a linear velocity of $0.2 m/s$ and an angular velocity of $0.2 rad/s$. $\xi_{des}$ shows the planned DCM trajectory and the footprints are visualized as polygons in green and orange.}
    \label{fig:dcm_foot_tracking}
\end{figure*}

\begin{figure}
    \centering
    \includegraphics[width=0.95\linewidth]{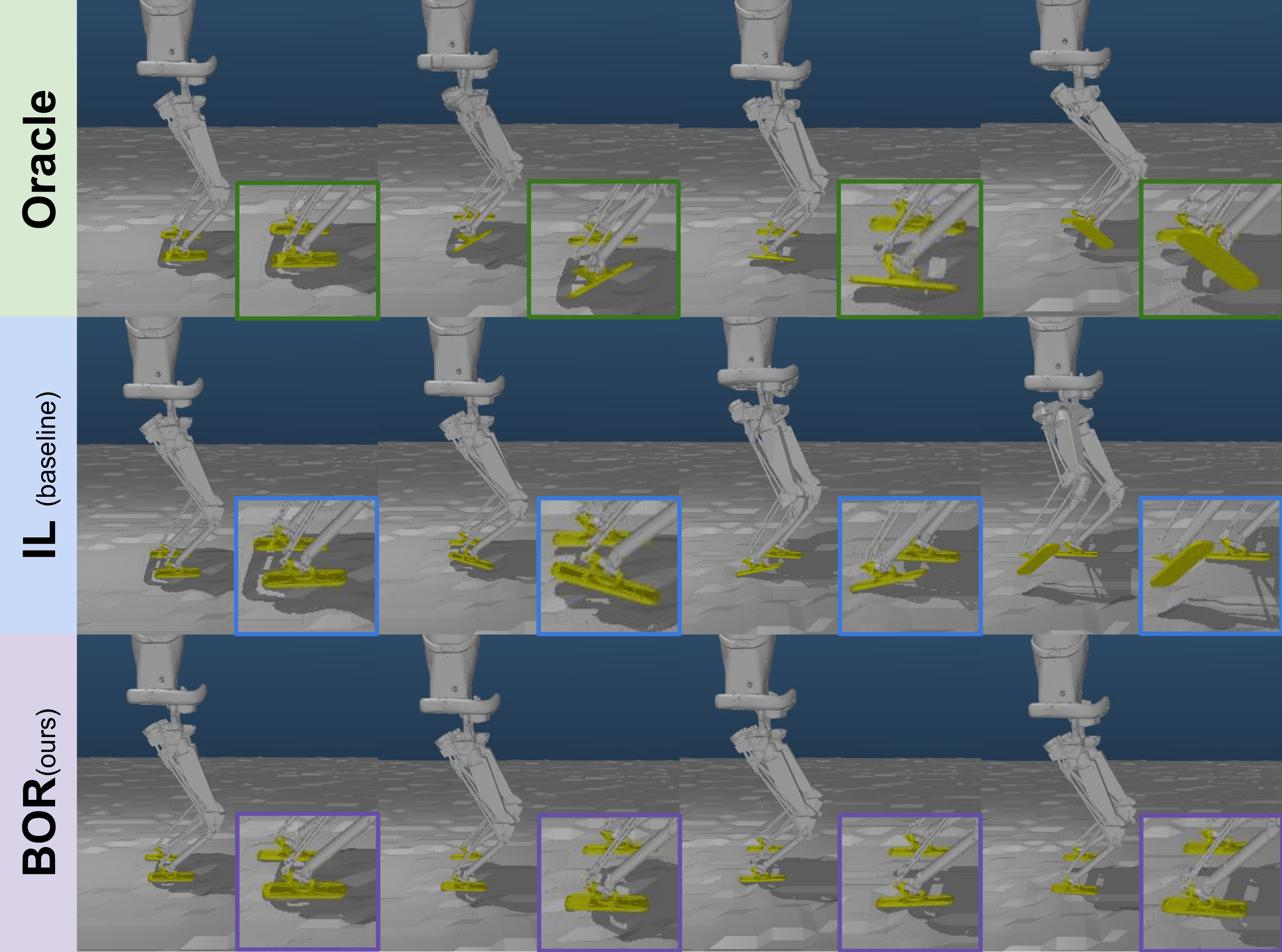}
    \caption{\textbf{Evaluation on uneven terrain.} Uneven terrain introduces additional uncertainties in real-world environments, yet even the Oracle policy struggles to explicitly encode ground irregularities. Thanks to the learning paradigms in our framework, our method learns adaptive behaviors that go beyond simply mimicking the Oracle, unlike the baseline IL approach.}
    \label{fig:uneven_terrain}
\end{figure}
\subsubsection{Comparison with Baselines}

To evaluate the effectiveness of our methods in improving robustness and performance for torque-based locomotion, we compare our approach against baseline methods under domain randomization, as detailed in Table \ref{tab:domain_randomization}. We validate the proposed key questions \textbf{Q1–Q4} and compare our methods against the Oracle and Base policies and the baselines in Sec. \ref{sec:baselines}. For each approach and robot, we conduct 10 simulations, where robots are tasked with following commanded linear and angular velocities. Each simulation covers 5 seconds of walking, and we vary the velocity commands every 2 seconds. We early terminate the simulations if robots are falling.

Table \ref{tab:comparison} summarizes the results across the metrics introduced in Sec. \ref{sec:metrics}. The Oracle policy (Algorithm \ref{alg:oracle}) achieves the best performance, successfully completing all experiments with minimal tracking errors in both DCM and foot trajectories. This demonstrates its capacity to effectively supervise RL agents during training (\textbf{Q1}). In contrast, the Base policy—using the same model-based controller as the Oracle but without knowledge of system uncertainties—fails to execute the tasks, highlighting its brittleness under randomization.

Our method (BOR), shown in the last row, successfully compensates for the Base policy's limitations. After training with Oracle supervision, the residual policy approaches Oracle-level performance, even slightly outperforming it in DCM tracking for Kangaroo and Return metric for Bruce (\textbf{Q2}). Regarding \textbf{Q3}, our supervised residual RL method significantly outperforms standard residual RL, which frequently fails due to the challenges of learning torque-based locomotion for high degree-of-freedom robots—consistent with prior findings \cite{Torque-Based, 10375154}. However, our results also show that the performance of pure imitation learning (IL) varies across robotic platforms. While IL performs comparably to BOR on Kangaroo, it performs significantly worse on Bruce. This suggests that IL is more sensitive to the Oracle policy’s reliability. For Bruce with line feet, although the Oracle has access to ground-truth system states, it sometimes results in unstable control during rapid transitions in commanded walking velocity. In such cases, RL helps explore corrective actions that deviate from the Oracle to improve stability and overall performance. Consequently, RL enhances both DCM tracking and foot tracking performance in BOR compared to IL. Finally, in response to \textbf{Q4}, our supervised residual RL framework consistently achieves near-Oracle performance across all three bipedal robots under the same training setup.

In addition, Table \ref{tab:noise_level} presents the impact of varying domain randomization levels on the performance of the model-based controller (MBC) and our proposed method (BOR). This analysis is conducted by adjusting the parameter $\beta$, as defined in Table \ref{tab:domain_randomization}. At a low uncertainty level ($\beta=0.1$), MBC achieves a $100\%$ success rate, demonstrating its robustness under mild system perturbations. However, its performance degrades significantly as uncertainty increases, with failure rates rising to $50\%$ at $\beta=0.5$ and $90\%$ at $\beta=0.7$. In contrast, BOR—trained with the highest level of domain randomization ($\beta=1$)—maintains a $100\%$ success rate and exhibits minimal DCM and foot tracking errors across the entire range $\beta \in [0, 1]$, highlighting its superior robustness to system uncertainties.

Furthermore, we evaluate the training progress of the Kangaroo robot in Fig. \ref{fig:training_procedure}. Thanks to the design of our Oracle policy and its supervision, our method efficiently learns the torque control, achieving over an 80\% success rate after just 1,500 episodes—equivalent to 4.16 hours of simulation time in a single environment. In contrast, standard residual RL shows little to no progress during training, primarily due to the limited design of our reward functions. Our reward structure is intentionally much simpler than in prior work \cite{rudin2022learning}, and unlike their approach—which leverages thousands of parallel environments—we train using only a single environment. Despite these constraints, our method successfully learns the desired performance by combining supervision from the Oracle and reward signals during training.

\subsubsection{Robustness on uneven terrains}
In addition to uncertainties in the robotic system, ground unevenness introduces another major source of uncertainty in locomotion tasks. Because it is challenging—or even infeasible—to generate optimal planning trajectories for variables such as foot placement and ankle touchdown angles on uneven terrain, these irregularities pose significant challenges for the Oracle in guiding the learning process. In this work, we always assume a flat floor for the Oracle policy while leveraging the learning paradigm to handle uncertainties. Figure \ref{fig:uneven_terrain} presents simulated results using identical walking parameters under different approaches. The Oracle policy leads to unstable control due to unmodeled ground unevenness, and this instability is directly inherited by the imitation learning (IL) baseline, since IL strictly mimics the Oracle’s behavior. In contrast, our method (BOR) enables adaptive behavior by allowing the policy to explore its action space through reinforcement learning. The simulation results demonstrate that our method can further optimize the policy beyond merely imitating the Oracle.

\subsubsection{Command tracking}
Leveraging the trajectory generators in our model-based pipeline, we obtain DCM and foot trajectories that enable the robots to follow commanded velocities effectively. To evaluate the velocity tracking performance, we plot the planned and measured trajectories for the H1-2 and Kangaroo robots in Figure \ref{fig:dcm_foot_tracking}. In these experiments, the H1-2 robot is commanded to walk forward at $0.2m/s$, while the Kangaroo receives an additional $0.2rad/s$ angular velocity.
Our method tracks the planned trajectories as smoothly and accurately as the Oracle policy, demonstrating the ability to closely follow the commanded velocities under domain randomization.

\subsubsection{Ablation study}
We conduct an ablation study to evaluate the role of residual learning in our framework. For that, we combine Oracle supervision with standard RL, in which our RL policy directly learns to command torques rather than residuals. Therefore, the Base policy is not used for training or for inference. As shown in Table \ref{tab:comparison} and Figure \ref{fig:training_procedure}, learning residual torques improves learning efficiency, but the overall performance gains are limited.

To further investigate this behavior, we analyze torque tracking for the Kangaroo robot's right hip joint, as illustrated in Figure \ref{fig:torque_tracking}. Due to the extensive domain randomization, the Base policy produces torque commands that deviate significantly from the desired targets, resulting in poor performance. Consequently, learning residuals on top of the Base policy offers limited benefits compared to directly learning the torque commands themselves.

Despite the large discrepancy between the Base and Oracle policies, our RL agents successfully learn to compensate for these differences, as evidenced by the improved tracking shown in Figure \ref{fig:torque_tracking}.

\section{Discussion}
Learning torque-level locomotion policies with reinforcement learning is typically data-inefficient, while model-based optimization methods rely on accurate system models that rarely hold in real-world settings. Our approach bridges these two paradigms by learning a residual torque policy guided by an Oracle while explicitly modeling real-world uncertainties in simulation.

To balance imitation and exploration, we jointly optimize reinforcement and supervision objectives. Oracle guidance significantly improves learning efficiency; however, even an Oracle with access to ground-truth states cannot anticipate all adverse conditions. For instance, abrupt user command changes or unstable ground contacts can cause failures that the Oracle cannot prevent. As shown in our results on the Bruce and H1 robots, over-reliance on the Oracle can therefore limit performance under highly stochastic conditions.

\section{Conclusion}
In this work, we present a supervised reinforcement learning framework for torque-based humanoid locomotion that addresses two fundamental challenges: the sim-to-real gap inherent in model-based control and the data inefficiency of standard model-free RL. To improve robustness to unknown system uncertainties, we employ domain randomization during training, allowing the learned policies to generalize across a wide range of simulated conditions. To further enhance learning efficiency, we introduce a model-based Oracle policy that provides informative supervision despite being suboptimal under randomized dynamics.

Oracle supervision is incorporated into the RL training process through an additional loss term that explicitly encourages the policy to align with the Oracle during gradient updates. This direct guidance significantly accelerates learning and leads to improved performance compared to standard RL methods that rely solely on reward signals.

We evaluate our framework on three bipedal robots—Kangaroo, Unitree H1-2, and Bruce—and show that the learned policies achieve performance comparable to their respective Oracles across all platforms. Notably, the same training framework is applied without robot-specific modifications. Ablation studies further demonstrate that our method effectively supports both residual torque learning on top of a base controller and direct torque command learning.

Overall, our results demonstrate that Oracle-guided supervised reinforcement learning provides an efficient, robust, and transferable solution for torque-based locomotion control in humanoid robots.




\bibliographystyle{IEEEtran}
\bibliography{references, ref}

\vfill

\end{document}